\documentclass[conference]{IEEEtran}
\IEEEoverridecommandlockouts
\usepackage{cite}
\usepackage{amsmath,amssymb,amsfonts}
\usepackage{algorithmic}
\usepackage{graphicx}
\usepackage{textcomp}
\usepackage{xcolor}
\usepackage[format=hang,labelfont=bf,font=small]{caption}
\usepackage{multirow}
\usepackage{tablefootnote}
\usepackage{subcaption}
\usepackage{enumitem}

\def\BibTeX{{\rm B\kern-.05em{\sc i\kern-.025em b}\kern-.08em
    T\kern-.1667em\lower.7ex\hbox{E}\kern-.125emX}}
\begin{document}

\title{Learning to search for and detect objects in foveal images using deep learning\\ 
}
\author{\IEEEauthorblockN{Beatriz Paula}
\IEEEauthorblockA{\textit{Instituto Superior Técnico}\\
\textit{Universidade de Lisboa}\\
Portugal \\
beatriz.paula@tecnico.ulisboa.pt}
\and
\IEEEauthorblockN{Plinio Moreno}
\IEEEauthorblockA{\textit{Insitituto de Sistemas e Robótica \& Instituto Superior Técnico} \\
\textit{Universidade de Lisboa}\\
Lisboa, Portugal \\
plinio@isr.tecnico.ulisboa.pt}
}

\maketitle
\thispagestyle{plain}
\pagestyle{plain}

\begin{abstract} The human visual system processes images with varied degrees of resolution, with the fovea, a small portion of the retina, capturing the highest acuity region, which gradually declines toward the field of view's periphery. However, the majority of existing object localization methods rely on images acquired by image sensors with space-invariant resolution, ignoring biological attention mechanisms. 

As a region of interest pooling, this study employs a fixation prediction model that emulates human objective-guided attention of searching for a given class in an image. The foveated pictures at each fixation point are then classified to determine whether the target is present or absent in the scene. Throughout this two-stage pipeline method, we investigate the varying results obtained by utilizing high-level or panoptic features and provide a ground-truth label function for fixation sequences that is smoother, considering in a better way the spatial structure of the problem. 

Finally, we present a novel dual task model capable of performing fixation prediction and detection simultaneously, allowing knowledge transfer between the two tasks. We conclude that, due to the complementary nature of both tasks, the training process benefited from the sharing of knowledge, resulting in an improvement in performance when compared to the previous approach's baseline scores.

\end{abstract}


\section{Introduction}

\par Computer Vision (CV) is an interdisciplinary area that connects several lines of study such as Computer Science, Biology, Psychology, Engineering, Mathematics and Physics. From an engineering standpoint, it attempts to understand and model the visual system in order to interpret the visual world \cite{b1}. It has many applications such as object detection, face recognition and action and activity recognition, in a multitude of fields from military to medicine.

\par A fundamental difference between the human visual system and current approaches to object search is the acuity of the image being processed \cite{b2}. The human eye captures an image with a very high resolution in the fovea, a small region of the retina, and a decrease in sampling resolution towards the periphery of the field of view. This biological mechanism is crucial for the real-time image processing of the rich visual data that reaches the eyes (0.1-1 Gbits), since visual attention prioritizes interesting and visually distinctive areas of the scene, known as salient regions, and directs the gaze of the eyes. In contrast, image sensors, by default, are designed to capture the world with equiresolution in a homogeneous space invariant lattice \cite{b3}, and current solutions to vision system performance rely on the increase of the number of pixels. This limits real-time applications due to the processing bottleneck and the excessive amount of energy needed by state-of-the-art technologies.

\par In the last decade, Deep Learning (DL) has shown tremendous success when compared to traditional Machine Learning (ML) approaches. First proposed in the 80s \cite{b4}, the Convolutional Neural Network (CNN), a DL technique, is inspired by the human visual processing system. In 2012, in the ImageNet Large Scale Visual Recognition Challenge, the winner, Alex Krizhevsky, introduced a CNN implementation solution \cite{b5} that showed its massive power as a training architecture. The AlexNet shows a lot of similarities with LeNet \cite{b6} that was published in 1998. However, by scaling up both the data and the computational power, we are only now seeing the true potential of Deep Learning in Computer Vision. 

\par Nevertheless, it remains challenging to replicate and model the human visual system. In an attempt to solve this, there have been advancements in combining these DL techniques with image foveation and saliency detection models. It has been shown that a "foveated object detector can approximate the performance of the object detector with homogeneous high spatial resolution processing while bringing significant computational cost savings" \cite{b7}.

\par The primary objective of this thesis is to utilize goal-guided scanpath data for object detection in images with foveated context. The following is a description of the problem we intend to solve: given an input image and an object category, indicate the presence or absence of instances of that class in the scene while adjusting the acuity resolution to mimic the human visual system. 

\par Our contributions include: (i) Benchmark of recent approaches based on Deep Learning, which are able to predict fixations, on a recent large-scale dataset; (ii) a ground-truth label function for fixation sequences that is smoother, considering in a better way the spatial structure of the problem; (iii) evaluation of two alternative visual representations (conventional high-level features from VGG and a more elaborate multi-class presence description); and (iv) the introduction of a novel dual task approach that simultaneously performs fixation and target detection.

\section{Related Work}

\par Human attention is driven by two major factors, bottom-up and top-down factors \cite{b8}. While bottom-up is driven by stimulus properties such as distinct features in the field of vision, which means saliency detection is executed during the pre-attentive stage, top-down factors are influenced by behaviourally relevant stimuli, such as prior knowledge, expectations and goals \cite{b9}. This latter factor can be detected in the 1967 experiment where individuals where asked to analyze a scene with different goals \cite{b10}. In this experiment individuals observed a picture of a family in a room with a visit, and were to answer different questions, such as to freely examine, to determine the people's ages, the material circumstances of the family, etc. According to Yarbus “depending on the task in which a person is engaged, i.e., depending on the character of the information which he must obtain, the distribution of the points of fixation on an object will vary correspondingly, because different items of information are usually localized in different parts of an object”.

\par The CV task of Gaze Prediction aims to predict fixation patterns made by people in image viewing and can have a spatial representation, in fixation density maps, and an added temporal representation when predicting scanpaths. In this area of study, most work focuses on free-viewing, which, as mentioned, is led by bottom-up attention. 

\par In the method proposed by Ngo and Manjunath \cite{b12}, CNNs are used for feature extraction and feature maps compilation which are then used in a Long Short Term Network (LSTM) responsible for modeling gaze sequences during free-viewing. This latter network was introduced in \cite{b13} as a solution to the vanishing gradient problem of RNNs that prevented the understanding of long dependencies since the correlation structure died down during the backward pass. These networks have a more complex structure which tweak the hidden states with an additive interaction, instead of a linear transformation previous recurrent networks performed, allowing the gradient to fully backpropagate all the way to the first iteration. 

\par However, as previously mentioned, human scanpaths during search tasks vary depending on the target items they are trying to gain information from, therefore guided search cannot be predicted based on free-viewing knowledge, where there were no explicit goals. Goal-directed attention is additionally relevant due to the human search efficiency in complex scenes that accounts for scene context and target spatial relations \cite{b14}. 

\par In \cite{b15}, a similar approach to the free-viewing scanpath predictor presented in \cite{b12} was taken. However, they leveraged a Convolutional Long Short Term Memory (ConvLSTM) architecture, and introduced a foveated context to the input images on top of an additional input encoding the search task, which found human fixation sequences to be a good foundation for object localization. The ConvLSTM had been previously introduced in \cite{b16} as a variant of LSTMs better suited for 3-dimensional inputs, such as images. This adaptation still contains the same two states: a hidden state, \(h\), and a  hidden cell state, \(c\); and the same four intermediate gates: the input gate \(i\), forget gate \(f\), output gate \(o\) and candidate input \(\Tilde{c}\); as the LSTM architecture. However, a convolution is performed during the computation of the gates instead of the previous product operations, as seen in the following equations:
\setlength{\belowdisplayskip}{1pt} \setlength{\belowdisplayshortskip}{1pt}
\setlength{\abovedisplayskip}{1pt} \setlength{\abovedisplayshortskip}{1pt}
\begin{equation}
\label{eq:lstmi}
    i_t = \sigma(W_i * x_t + U_i * h_{t-1})
\end{equation}
\begin{equation}
\label{eq:lstmf}
    f_t = \sigma(W_f * x_t + U_f * h_{t-1})
\end{equation}
\begin{equation}
\label{eq:lstmo}
    o_t = \sigma(W_o * x_t + U_o * h_{t-1})
\end{equation}
\begin{equation}
\label{eq:lstmictilde}
   \Tilde{c}_t = \tanh{(W_c * x_t + U_c * h_{t-1})}
\end{equation}
\begin{equation}
\label{eq:lstmc}
   c_t = f_t \odot c_{t-1} + i_t \odot \Tilde{c}_t
\end{equation}
\begin{equation}
\label{eq:lstmh}
   h_t = o_t \odot tanh(c_t)
\end{equation}

\noindent where $\odot$ denotes an element wise product, $*$ denotes a convolution, and \(W\) and \(U\) are the weight matrices of each gate that operate over the hidden states.

\par The limited amount of available data containing human scanpaths in visual search was, however, identified as a significant obstacle in \cite{b15}. Since then a new large-scale dataset has been introduced in \cite{b17}, which has shown promising results in \cite{b18}, where an inverse reinforcement learning algorithm was able to detect target objects by predicting both the action (fixation point selection) and state representations at each time step, therefore replicating the human attention transition state during scanpaths. This approach additionally utilized features extracted from a Panoptic FPN model \cite{b20}, that performs panoptic segmentation which is the unification of "the typically distinct tasks of semantic segmentation (assign a class label to each pixel) and instance segmentation (detect and segment each object instance)" \cite{b19}. 

\par Lastly, recent works in CV have begun experimenting with a Transformer-styled design, as it has been demonstrated that such architectures offer top performance in the field of Natural Language Processing \cite{b22}. For example, visual attention-driven transformers have been applied to the medical field to aid in diagnostics, as seen in \cite{b24} and \cite{b25} where scanpath data provided task-guided visual attention that aided in disease classification in histopathology images and chest radiographs, respectively.

\section{Systems Overview}

\par In this section, we present the architecture of the two strategies used in this study: a two-stage pipeline system consisting of a gaze fixations predictor and an image classifier, and a dual-task model that conducts scanpath prediction and target detection simultaneously.

\subsection{Fixation Prediction Module}

\par We undertake tests on high-level and panoptic image feature inputs within the fixation prediction module. Despite configuring a unique architecture for each model, they both share the same structure. At each time-step $T=t$, the Input Transformation Section aggregates the features of the foveated pictures at each fixation location since the beginning of the gaze sequence, $T \in 0,..., t$, as well as the task encoding of the target object. This combined input is then sent to the Recurrent Section, which uses ConvLSTM layers to emulate human-attention through its hidden states. The Recurrent Section then outputs its final hidden state $h_{t+1}$ to the model's Output Section, which predicts the next scanpath fixation as a discrete location in an image grid with dimensions $H \times W$. We will now present the architecture of each model in detail. 

\subsubsection{Fixation Prediction from High-Level Features} 
\label{subsub:FP1}

\par In this model, we utilized the high-level features retrieved from the ImageNet-trained VGG16 model \cite{b26, b27} with dimensions $H \times W \times Ch$, and it is composed by the following sections:

\begin{itemize}
    \item \textbf{Input Transformation}: To aggregate the feature maps and the task encoding, we opted for a Multiply layer which executes the element-wise multiplication of these inputs. In addition, depending on its format, the task encoding may be transmitted through a Fully Connected (FC) Layer with $Ch$ units and a tanh activation, followed by a Dropout Layer with a rate of $r_{Dropout}$ in order to prevent overfitting.
    
    \item \textbf{Recurrent Section}: This portion mainly consists on a ConvLSTM layer composed of $F$ filters, which correspond to the dimensionality of its output, a kernel size of $K$ x $K$, a stride of $S$, and a left and right padding of $P$. This layer has a tanh activation, and the recurrent step utilizes a hard sigmoid activation, a piece-wise linear approximation of the sigmoid function, for faster computation. Subsequently, to prevent overfitting we perform batch normalization, where we normalize the vector $h$ with the batch mean $\mu$ and batch variance $\sigma$ of the current batch during training, as seen in equation \ref{eq:BNTrain}, where $\epsilon$ is a small constant, and $\alpha$ and $\beta$ are learned scaling and offset factors, respectively. During inference, the arrays are normalized with a moving mean $M$ and moving variance $V$, which are non-trainable variables that were determined throughout training with equation \ref{eq:BNmeantest}, where $\mu$ and $\sigma$ are the mean and variance of the test batch and $\gamma$ is the momentum hyper-parameter.
    \vspace{5pt}
    \begin{equation}
    \label{eq:BNTrain}
        y = \alpha \cdot \frac{x - \mu}{\sqrt{\sigma + \epsilon}} + \beta
    \end{equation}
    \begin{equation}
    \label{eq:BNmeantest}
        M = M \cdot \gamma + \mu \cdot (1-\gamma) \qquad  V = V \cdot \gamma + \sigma \cdot (1-\gamma)
    \end{equation}
    \vspace{1pt}
    \item \textbf{Output Section}: We first perform a flattening operation to each temporal slice of the input with the help of a Time Distributed wrapper. The flattened array is then fed to a FC layer and has $H \times W$ units and a softmax activation function.
\end{itemize}
\smallskip

\subsubsection{Fixation Prediction from Panoptic Features} 

\par To compute these new features we resorted to the Panoptic FPN model, introduced in \cite{b20}. Due to the higher inference time of this model, of 53.0 ms/image, when compared to the VGG16, of 4.2 ms/image, with a GPU, we pre-computed the belief maps for each image at a high resolution, with the original image at full acuity, and a low resolution, after applying a uniform Gaussian blur filter with a blur radius of $\sigma$. This way we simulate the human foveal system by computing the belief map $B$ as:
\begin{equation}
\label{eq:panopBelief}
    B(t) = M_t \odot H + (1 - M_t) \odot L
\end{equation}
where $H$ and $L$ are the belief maps for the high and low resolution images, respectively, and $M_t$ is the binary fixation mask at time step $t$, of size $H \times W$, where every element is set to $0$ except the grid cells at an euclidean distance shorter than $r$ from the current fixation point.

\par To duplicate the belief maps in \cite{b18}, the task encoding is now a one-hot encoding with dimensions $H \times W \times Cl$, where each row of the axis $Cl$ corresponds to an object class and the two-dimensional map $H \times W$ is all set to one for the target class and zero for the others. This input is subsequently transmitted to the Input Transformation stage, where it is concatenated with the image feature maps.

\par The recurrent section of the model is composed of $d$ ConvLSTM layers, each followed by a Batch Normalization layer. Every ConvLSTM is constructed with the same hyper-parameters: each one has $F_{LSTM}$ filters with a kernel size of $K_{LSTM} \times K_{LSTM}$, a stride of $S_{LSTM}$, a padding of $P_{LSTM}$, a tanh activation function and a hard sigmoid activation during the recurrent step.

\par Finally, in the output section, we conducted experiments over two different setups. The first one is composed of a 3d-Convolutional layer with a sigmoid activation followed by a time distributed flattening operation. The second setup is comprised of the same 3d-Convolutional layer, but with a ReLu activation, and a flattening layer followed by a Fully Connected layer with softmax activation.

\subsection{Target Detection Module}
\label{sub:det}

\par In the last stage of the pipeline, the model finally gets to evaluate, at each time-step of the sequence, if the fixation point coincides with the location of the target object. To achieve this, we chose to once again leverage a state of the art classifier, the VGG16, and develop 18 binary classifiers, one for each task. In addition, as a baseline, we utilize a complete VGG16 trained on the ImageNet dataset to perform classification on our data, replicating an approach taken in \cite{b29}.

\par In order to fine-tune the already pre-trained VGG16 model, we substituted its classification layers, after the last convolutional layer, with three fully connected layers, $FC_i$ with $i \in \{1, 2, 3\}$, each with $U_i$ units, where $FC_1$ and $FC_2$ were followed by a ReLu activation function and $FC_3$ was followed by a sigmoid action function. During training, only the parameters of these last $FC$ layers were updated. 

\subsection{Dual Task Model}

\par
\begin{figure}[h]
    \centering
    \begin{subfigure}[b]{0.85\linewidth}
        \includegraphics[width=\textwidth]{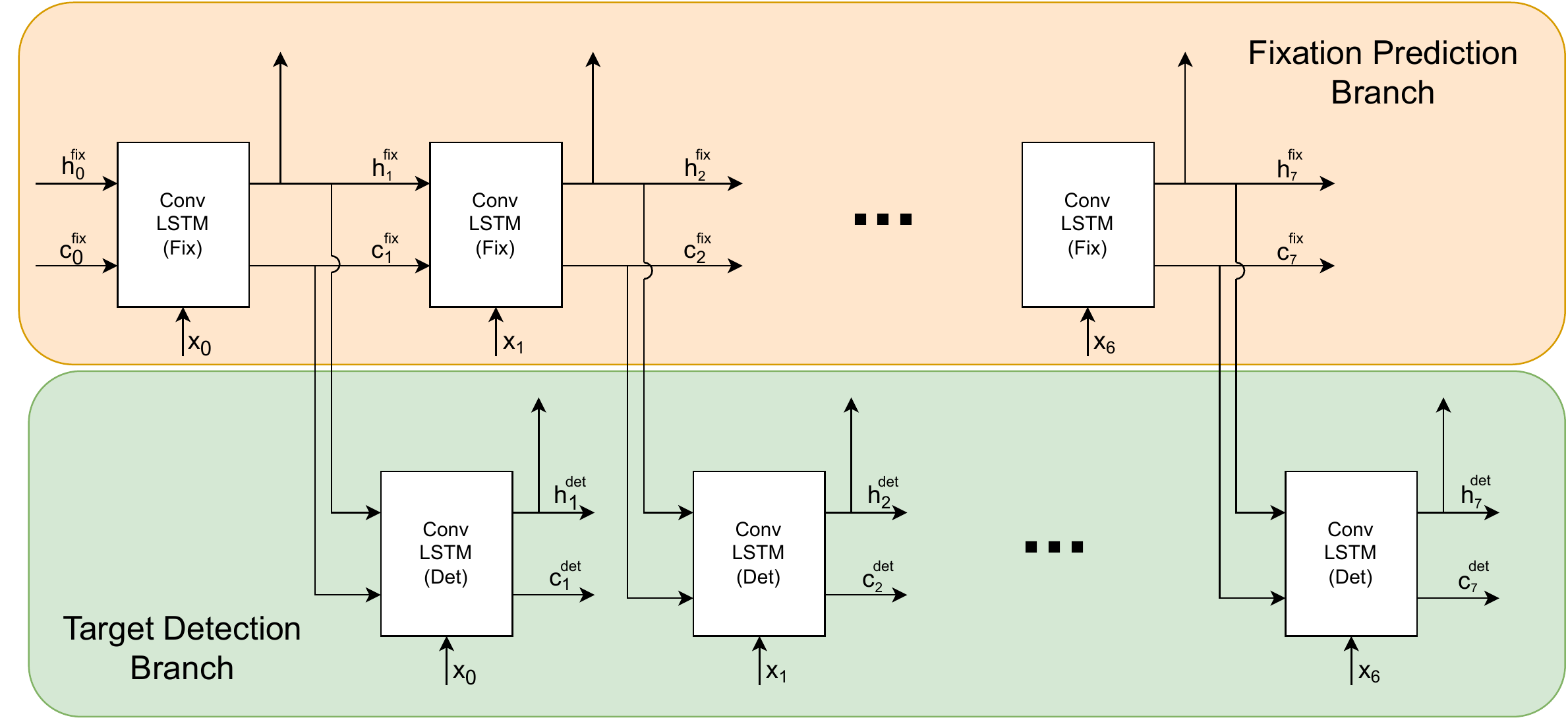}%
        \caption{Architecture A.}
        \label{subfig:dual_arch_A}%
    \end{subfigure}
    \vskip\baselineskip
    \begin{subfigure}[b]{0.85\linewidth}
        \includegraphics[width=\textwidth]{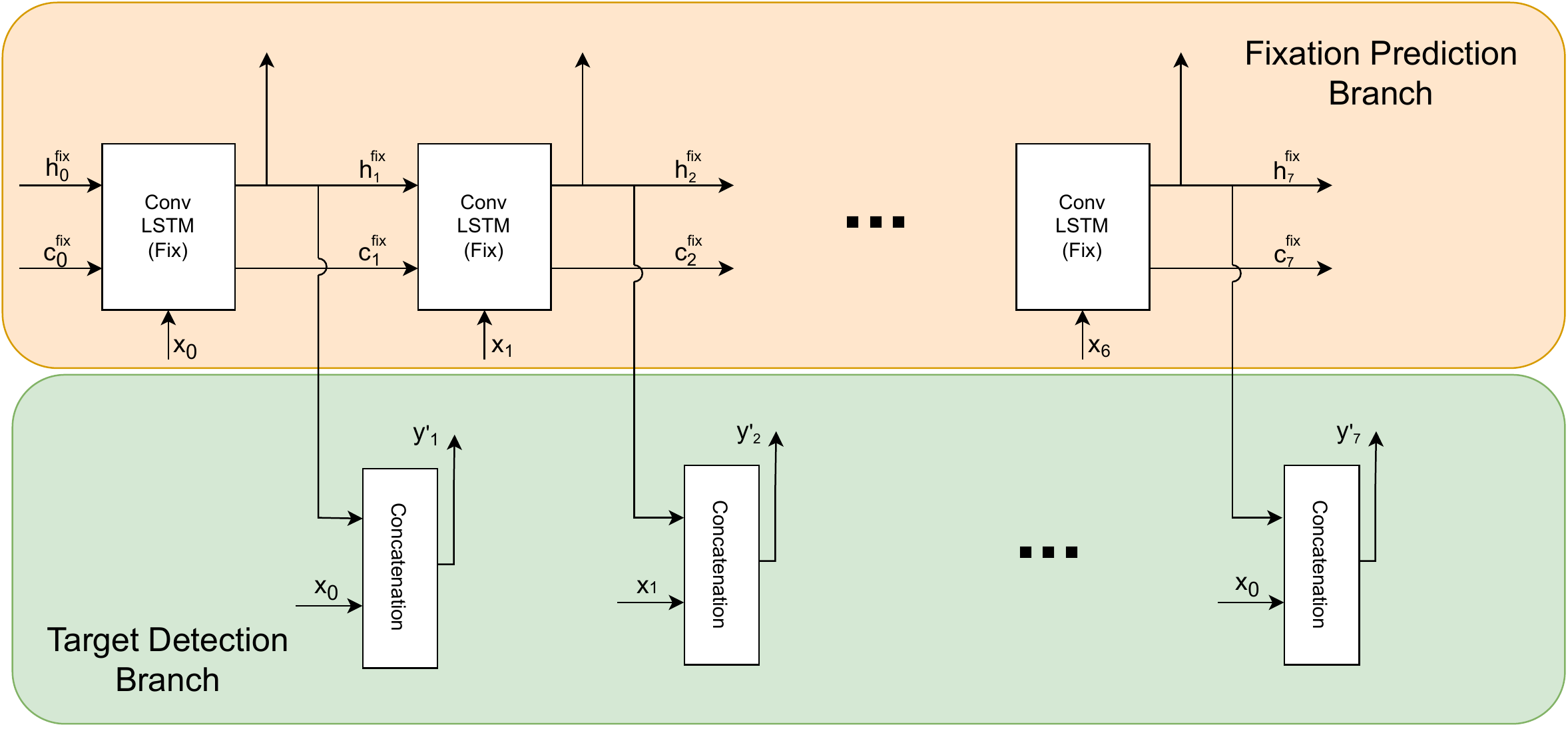}%
        \caption{Architecture C.}
        \label{subfig:dual_arch_C}%
    \end{subfigure}
    \caption{Information flow across the fixation prediction and target detection branch in architectures A, on the top, and C, on the bottom.}
    \label{fig:dual_archs}
\end{figure}

 Our multitasking approach to the problem of attention-guided object localization  considers three different architectural alternatives. The primary distinction between them is the employment of a ConvLSTM layer in both task branches in the first two models, but only in the fixation prediction branch in the third. All models receive as input the high-level feature maps, with dimensions $H \times W \times Ch$ and a one-hot task encoding array, of size $Cl$, which are then aggregated. Similar to the subsection \ref{subsub:FP1}, the grouping of both of these inputs is accomplished by passing the task encoding through a fully connected layer with $Ch$ units and tanh activation, and conducting an element-wise multiplication with the foveated image's feature maps. After this shared module, the models branch off to complete each specific task using the following architectures:

\begin{itemize}[leftmargin=*]
    \item \textbf{Architecture A} After performing the input transformation, where we aggregate the feature maps and task encoding, the array $x_t$ is fed to two ConvLSTM layers, each containing $F$ filters of size $K \times K$ with a stride of $S$, and an output tanh activation and a hard sigmoid activation in the recurrent step. However, following each iteration of the fixation prediction recurrent module, its internal states $h_t^{fix}$ and $c_t^{fix}$ are passed to the detection branch as the internal states, $h_{t-1}^{det}$ and $c_{t-1}^{det}$, of the preceding time step, as illustrated in figure \ref{subfig:dual_arch_A}. Subsequently, the compositions of the fixation prediction and target detection branches resemble those described in sections \ref{subsub:FP1} and \ref{sub:det}, respectively. In the first branch, a temporal flattening operation is performed to $h^{fix}$, followed by an output layer consisting of a FC layer with softmax activation. In the second we classify each temporal slice of $h^{det}$ by employing the same structure of three FC layers $FC_i \in \{1,2,3\}$, with $U_i$ units, respectively, where the first two layers have a ReLu activation while the output layer has a sigmoid activation.
    
    \item \textbf{Architecture B} Similar to the previous architecture, each task branch employs ConvLSTM layer. The sole difference is that we now conduct the iterations of the detection module first, and send the internal states $h_t^{det}$ and $c_t^{det}$ to the fixation prediction module for the preceding time steps $t-1$ 
    
    \item \textbf{Architecture C} This design contains a fixation prediction branch that is identical to that of architecture A. Nevertheless, the target detection branch no longer has a ConvLSTM layer. Instead, at each time step $t$, the combined input $x_t$ computed by the shared module is concatenated with the output of the ConvLSTM layer, $h_{t+1}^{fix}$ of the fixation prediction task, as illustrated in figure \ref{subfig:dual_arch_C}. Finally, this concatenation is followed by the same three FC layers utilized by the previous architectures.
\end{itemize}

\section{Implementation}

\par In this section we present the dataset, in addition to the conditions in which our models were trained and tested.

\subsection{Dataset}

\par The dataset utilized in this study was the COCO-Search18 dataset, introduced in \cite{b17}. In terms of the amount of images, target classes, and fixations, it remains the largest dataset on goal-directed behavior, to our knowledge. This dataset consists of 6,202 images from the Microsoft COCO dataset \cite{b30}, evenly split between target-present and target-absent images, of 18 target categories (bottle, bowl, car, chair, analogue clock, cup, fork, keyboard, knife, laptop, microwave, mouse, oven, potted plant, sink, stop sign, toilet and tv), with eye movement recordings from 10 individuals.

\par
\begin{figure}
    \centering
    \includegraphics[width=0.7\linewidth]{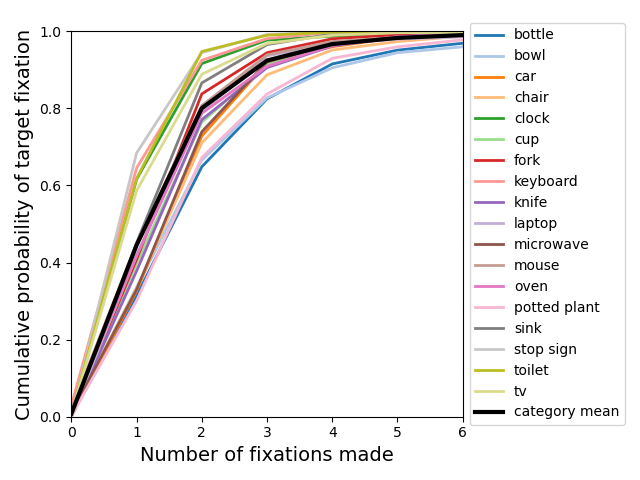}
    \caption{Cumulative probability of target fixation throughout sequences of the COCO-Search18 dataset train split}
    \label{fig:comulProbDataset}
\end{figure}

 As humans were able to fixate the target object with their gaze within their first six saccades 99\% of the time, as illustrated in figure \ref{fig:comulProbDataset}, fixation sequences with length greater than that were discarded. Additionally, the sequences were padded with a repeated value of the last fixation point to achieve a fixed length of 7, including the initial center fixation. This was done to replicate the procedure of a similar work \cite{b29}, where participants were instructed to fixate their gaze on the target object, once they found them, during search tasks.

\par To train the fixation prediction module and the dual task model, we used a random dataset split of 70\% train, 10\% validate and 20\% test over each class category and all images were resized to 320 x 512 which resulted in feature maps with $10 \times 16$ spacial dimensions. For the fine-tuning of the binary classifiers in the target detection module, due to the cluttered nature of the images in the dataset, we utilize a cropped region of the images surrounding each fixation point, with its size proportional to the median dimensions within each class.

\subsection{Training}

\par During the training phase, all our models were optimized with the Adam algorithm \cite{b33} and a learning rate of $lr = 0.001$, for a maximum of 100 epoch with an early stopping mechanism activated when the validation loss stops improving after a duration of 5 epochs. Additionally, every dropout is performed with $r_{Dropout} = 0.5$, the batch normalization uses $\epsilon = 0.001$ and $\gamma = 0.99$, and we use a batch size of 256 in every module apart from the fixation prediction performed with high-level features.

\begin{itemize}[leftmargin=*]
    \item \textbf{Fixation Prediction from High-Level Features} During training, we estimate the best weight and bias parameters that minimize the loss between the predicted output $\hat{y}$ and the ground truth label $y$, with the cross entropy function computed for every fixation time step $t$ for every sequence $s$ of each mini-batch $b$:
    
    \begin{equation}
    \label{eq:FPLoss}
        L_{CE} = - \sum_{s = 1}^{S} \sum_{t = 0}^{T} \sum_{i = 1}^{H \times W} y_i * log(\hat{y}_i)
    \end{equation}
    \smallskip
    \par \noindent where $S$ corresponds to the batch size, $T$ corresponds to the sequence length which is set to 6 (in addition to the initial fixation point at $t=0$) and $H \times W$ is the output size which is set to 160.
    
    \par We set with $F=5$ filters, a kernel size of $K=4$ and a stride of $S=2$, and, during training, we varied the batch size between 32, 64, 128 and 256 and we conducted an ablation study over theses additional hyper-parameters and settings:

    \begin{itemize}
    
    \item Fovea size: In this work we utilized the same real-time foveation system as in \cite{b15}, and assessed the model performance when varying the fovea size, which defines the radius of the region with highest visual acuity with the values of 50, 75 and 100 pixels.
    
    \item Task encoding: In order to encode the object class being searched, we experimented with several representations. The first was a one-hot encoding array of size 18 representing each class. The second was a normalized heat map of fixations made during the observations of that same task, compiled exclusively with training data. This encoding was represented both with a two dimensional array of size 10 x 16 and a one dimensional array of the flattened grid with size 160. When utilizing a one dimensional encoding, this array was passed through a FC layer with 512 units during the Input Transformation, while the two dimensional encoding was fed directly into the Multiply layer. 
    
     \item Ground truth: We consider both a one-hot encoding representation of the ground-truth label and a two dimensional Gaussian function with the mean set to the cell coordinates of the actual fixation location and the variance set to 1.
     
    \end{itemize}
    \smallskip
    
    \item \textbf{Fixation Prediction from Panoptic Features} During the training of this setup of the fixation prediction module we once again aim to minimize the loss function in equation \ref{eq:FPLoss}. Additionally, the feature maps used have dimension $10 \times 16 \times 134$ in order to replicate the scale of our grid shaped output. 
    
    \par In regards to the model architecture, we fixed the ConvLSTM layers' configuration to $F_{LSTM} = 10$ filters with square kernels of size $K_{LSTM} = 3$, a stride of $S_{LSTM} = 1$ and a padding of $P_{LSTM} = 1$ to maintain the features spatial resolution. In the output section, we set the $F_{Conv} = 1$ filter of the 3d-Convolutional layer to have a kernel size of $K_{Conv} = 2$, a stride of $S_{Conv} = 1$ and padding of  $P_{Conv} = 1$. The second setup of this section is configured to have a Fully Connected layer with 160 units.
    
    \par For this model, we additionally varied the depth of the recurrent section with \textbf{$d \in \{1, 3, 5\}$}, and altered the structure of the output section to utilize both a sigmoid and softmax as the final activation function. Concerning the data representation, we once again evaluated the impact of having a one-hot or a Gaussian ground truth encoding, and explored several belief maps settings: we varied the radius $r$ of the mask, $M_t$, with values $r \in \{1, 2, 3\}$ (each to emulate a corresponding fovea size of 50, 75 and 100); and experimented with a cumulative mask configuration, $M'_t$, where the binary mask utilized in equation \ref{eq:panopBelief}, in addition to the information of the current time step, accumulates the high acuity knowledge of all previous time steps. All panoptic feature maps were computed with a low resolution map $L$ extracted from a blurred input image with a Gaussian filter with radius $\sigma = 2$.
    \smallskip
    
    \item \textbf{Target Detection} The binary classifiers were implemented with each fully connected layer having $U_1=512$, $U_2=256$ and $U_3=1$ units, a dropout rate of $r_{Dropout}=0.5$, and we varied the fovea size between 50, 75 and 100 pixels. They were trained with a loss function defined as:

    \begin{equation}
    \label{eq:DetLoss}
        L_{BCE} = - \frac{1}{N}  \sum_{i=1}^N y_i\cdot log(\hat{y}_i) + (1- y_i) \cdot log(1 - \hat{y}_i)
    \end{equation}
    \smallskip
    \item \textbf{Dual Task} Concerning the architecture, the ConvLSTM layers are configured with $F=5$ filters of size $K=4$ to execute the convolutional operations with stride $S=2$, while the fully connected layers of the detection branch are configured with $U_1=64$, $U_2=32$ and $U_3=1$ units.
    \par During optimization we aim to minimize the combined loss $L_{Dual}$ defined as:
    
    \begin{equation}
        L_{Dual} = w_{fix} \cdot L_{fix} + (1- w_{fix}) \cdot L_{det}
    \label{eq:dualLoss_total}
    \end{equation}
    \smallskip
    
    where $L_{fix}$ and $L_{det}$ correspond to the loss of the fixation and detection prediction, respectively. The first is calculated using the categorical cross entropy function stated in equation \ref{eq:FPLoss}, while the second is calculated using the weighted binary cross entropy defined in equation \ref{eq:dualLoss_det}. 

    \begin{equation}
        w = y \cdot w_1 + (1-y) \cdot w_0
    \label{eq:dualLoss_weight}
    \end{equation} 
    \begin{equation}
        L_{det} = w \cdot [y * log(\hat{y}) + (1-y)\cdot log(1-\hat{y})]
    \label{eq:dualLoss_det}
    \end{equation}
    \smallskip

\par This final decision was made to address the imbalance in the detection data, as the target is absent in half of the images, and the detection is positive in a limited section of the scanpath sequences in the remaining images. To prevent the model from biasing its prediction on the predominant class, we set the relevance of the loss computed on the positive and negative detections with the weights $w_1$ and $w_0$ set to $1.6$ and $0.7$, respectively. We attained these values by, for the case of $w_1$ computing the multiplicative inverse of the ratio of positive detections on the total number of detections, and dividing it by 2; and, similarly, in the case of $w_O$ we perform the same computations, but with the ration of negative detections.

\par To determine the optimal configuration for each model's architecture, an ablation study was done over the fovea size and the degree of importance $w_{fix}$ of the fixation loss on the total loss. The former variable had its values set to 50, 75 and 100 pixels, while the latter had its values set to 0.10, 0.25, 0.50, 0.75, 0.90. 
\end{itemize}

\subsection{Prediction}

\par During the testing phase, both the fixation prediction module of the two-stage pipeline and the dual task model were used to predict a scanpath sequence of fixed length $l=7$, based on the training data, and we set the fixation point at $t=0$ as the center cell of the discretized grid. Additionally, they were both deployed with a beam search algorithm that selects the best $m$ fixation points at each time step, where $m$ is the beam width hyper-parameter which we set to 20, and appends them to the sequences they were generated from, while saving the target detection prediction in the case of the dual task model. In the next time step, the model runs for each of these $m$ predicted sequences, to select the next best $m$ predictions. This algorithm selects the optimal solution when considering the context given by the previously saved sequences, while still having a linear complexity of $\mathcal{O}(l) = m \times l$.

\par In regards to the target presence detector, all models were deployed on the scanpaths produced by the highest performing scanpath predictor. The baseline classifier was tested similarly to the binary models, except that the input cropped images are now constrained to a consistent size of $224\times224$. Due to the differences between the datasets, we adapt the ImageNet classes to our targets by grouping some sibling sub-classes (e.g. the task bowl corresponds to the union of the sub-classes mixing bowl and soup bowl), and we do not classify the tasks fork, sink, and oven because they do not exist in the ImageNet dataset. We classify a target as present when the corresponding class has the highest classification score and as absent otherwise.

\section{Results}
\subsection{Two Stage-Pipeline}
\subsubsection{Fixation Prediction}
\par
\begin{figure}
    \centering
    \includegraphics[width=0.85\linewidth]{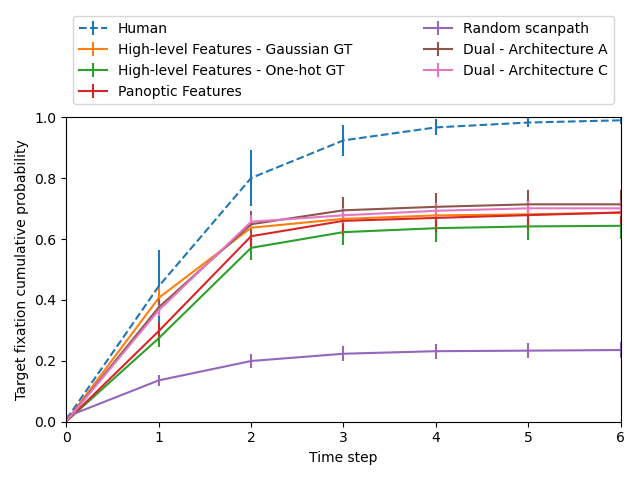}
    \caption{Search accuracy achieved by each model throughout scanpaths, with means and standard errors first computed over target classes.}
    \label{fig:TFP}
\end{figure}

 Regarding the fixation prediction task, we evaluate our models on several metrics. The first is \textbf{Search Accuracy} which is computed as the ratio of sequences in which a fixation point selects a grid cell that intersects the target's bounding box. The second is \textbf{Target Fixation Cumulative Probability (TFP)}, which is plotted in figure \ref{fig:TFP}, and presents the search accuracy attained by each time step. From this plot we can additionally attain the \textbf{TFP - Area Under Curve (TFP-AUC)} and the \textbf{Probability Mismatch}. The former computes the area under the TFP curve and the latter is the sum of absolute differences between the model's TFP and the human's observable data. Finally, we additionally compute the \textbf{Scanpath Ratio} as the ratio between the sum of euclidean distances between each fixation point and the distance from the initial fixation to the center of the target's bounding box. 

\par
\begin{figure}
\centering
    \begin{subfigure}[b]{0.4\linewidth}
        \includegraphics[width=\textwidth]{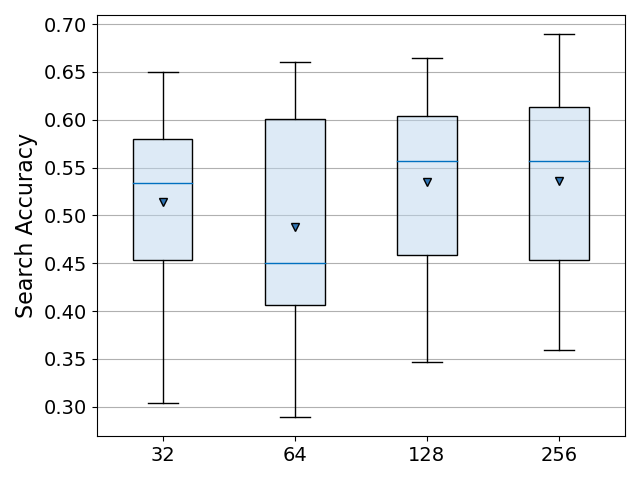}%
        \caption{Batch Size}
        \label{subfig:fp1_batch}%
    \end{subfigure}
    \qquad
    \begin{subfigure}[b]{0.4\linewidth}
        \includegraphics[width=\textwidth]{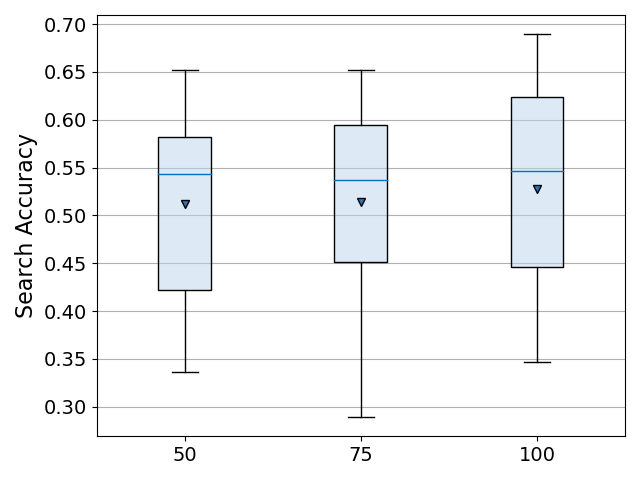}%
        \caption{Fovea Size}
        \label{subfig:fp1_fovea}%
    \end{subfigure}
    \vskip\baselineskip
    \centering
    \begin{subfigure}[b]{0.4\linewidth}
        \includegraphics[width=\textwidth]{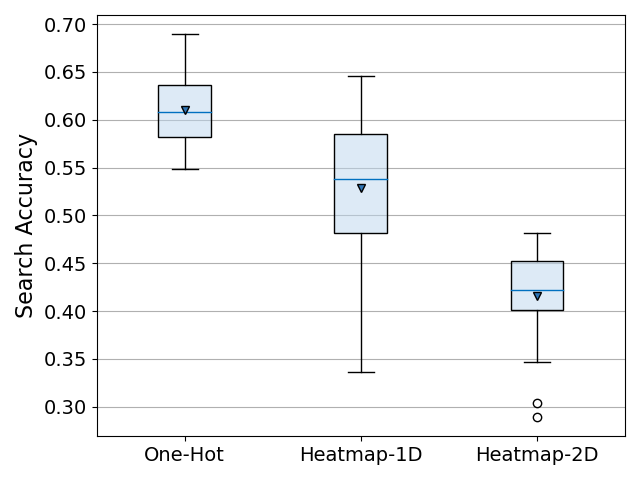}%
        \caption{Task Encoding}
        \label{subfig:fp1_task}%
    \end{subfigure}
    \qquad
    \begin{subfigure}[b]{0.4\linewidth}
        \includegraphics[width=\textwidth]{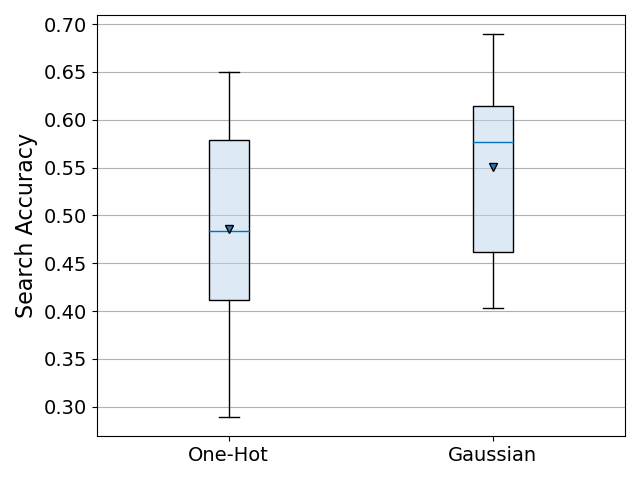}%
        \caption{Ground Truth}
        \label{subfig:fp1_gt}%
    \end{subfigure}
    
    \caption{Search accuracy box-plots of the single task High-level features' models, grouped by their training settings, with the mean values and outliers depicted as triangles and circles, respectively.}
    \label{fig:vgg_boxplots_configs}

\end{figure}

 Through the ablation study we conducted, we found that the high-level features scanpath predictor from the two stage pipeline was able to achieve higher search accuracy scores when using a one-hot task encoding and a Gaussian ground-truth, as seen in figure \ref{fig:vgg_boxplots_configs}, where the search scores are depicted in box-plots grouped by each training setting. This new label representation was able to outperform, with a search accuracy of 0.690, the traditional one-hot encoding that attained a score of 0.650, bench-marking the model presented in \cite{b29} on the COCO-Search18 dataset, which, previously, had only been able to achieve a score of 0.621.

\par In figure \ref{fig:TFP} we have plotted the TFP curve of several highest performing model configurations, in addition to the human search behavior\footnote{\label{note1}The Human TFP refers to the human behavior observed in the training data set due to the fact that the search fixations during the test dataset were withheld in the COCO-Search18. All the remaining performances depicted were computed for the test data split.}, and a random scanpath baseline model, which selected a random human sequence from the train data split for the same search target class. We observed a decrease in performance throughout time for all models through the slopes of each function. Apart from the randomized generator, the models were able to detect the majority of the targets by the second fixation step. Our models, however, reach a performance plateau much sooner than the human behavior, and barely detected any new targets throughout the last four fixation points. 

\par
\begin{figure}
    \centering
    \includegraphics[width=0.7\linewidth]{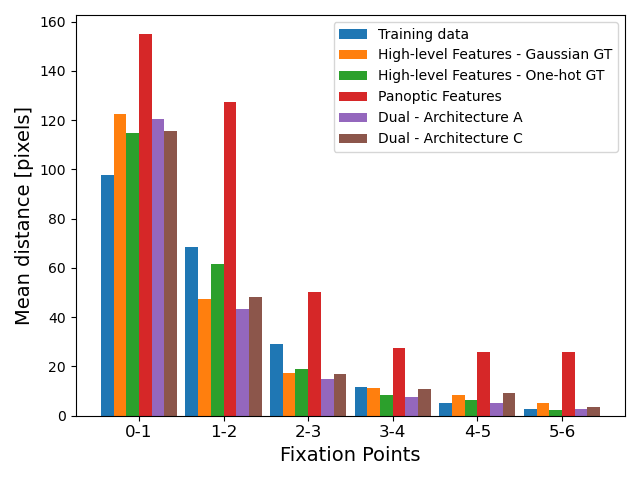}
    \caption{Euclidean distances between fixation points.}
    \label{fig:VGG_distances_fp}
\end{figure}

 With the intent to better understand these low performance ceilings, we first theorized that our models stopped searching after the second fixation point, and throughout the remainder of the scanpath predicted solely the neighboring grid cells. As depicted in figure \ref{fig:VGG_distances_fp}, the predicted sequences do converge throughout the time steps, however, this particularity is simply a replica of the training data behavior. 

\par When evaluating our model against two models presented in \cite{b18}, which also use panoptic features, the IRL and the BC-LSTM, we found that our models outperform the latter across all criteria, although the former approach seems to imitate human task-guided attention more accurately. The IRL utilizes inverse reinforcement learning, while the BC-LSTM utilizes the same ConvLSTM structure as our work, but instead of using it to directly predict the scanpaths they use it in an intermediate step to update the feature maps. 

\begin{figure}
    \centering
    \begin{subfigure}[b]{0.395\linewidth}
        \includegraphics[width=\textwidth]{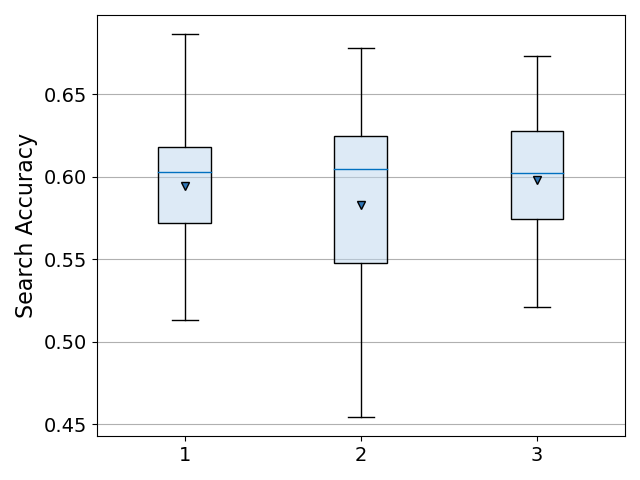}%
        \caption{Mask Radius.}
        \label{subfig:fp2_radius}%
    \end{subfigure}
    \qquad
    \begin{subfigure}[b]{0.395\linewidth}
        \includegraphics[width=\textwidth]{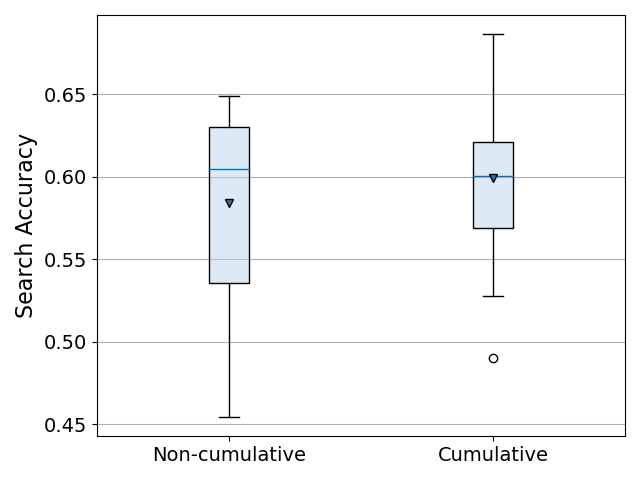}%
        \caption{Mask configuration.}
        \label{subfig:fp2_cumul}%
    \end{subfigure}
    \vskip\baselineskip
    \begin{subfigure}[b]{0.395\linewidth}
        \includegraphics[width=\textwidth]{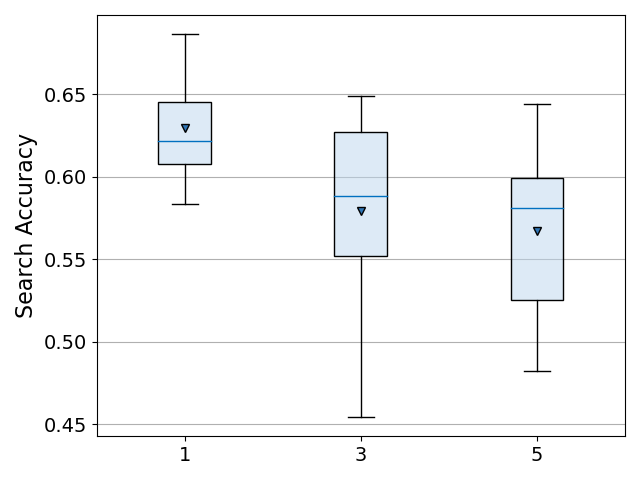}%
        \caption{Depth of recurrent section.}
        \label{subfig:fp2_depth}%
    \end{subfigure}
    \qquad
    \begin{subfigure}[b]{0.395\linewidth}
        \includegraphics[width=\textwidth]{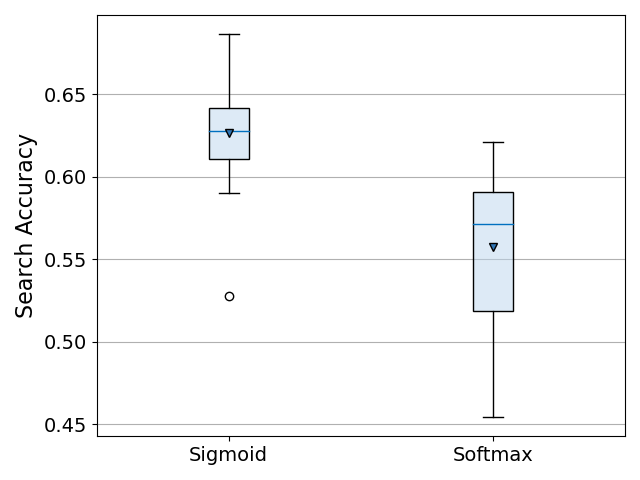}%
        \caption{Activation Function.}
        \label{subfig:fp2_activation}%
    \end{subfigure}
    \caption{Search accuracy box-plots of the single task Panoptic features' models grouped by their training settings, with the mean values and outliers depicted as triangles and circles, respectively.}
    \label{fig:panop_boxplots_configs}

\end{figure}

\par In turn, the panoptic features' fixation prediction model was found, through figure \ref{fig:panop_boxplots_configs}, to reach higher accuracies with a single ConvLSTM layer in its recurrent section, indicating an overfitting of the model as the number of network parameters increases. In addition, while the highest search accuracy score of 0.686 was attained with a cumulative mask of radius $r=1$, we could not detect a direct correlation between these settings and the model's performance.

\par Surprisingly, we can also observe that using a final sigmoid activation function greatly improved the model's results, despite the fact that this activation function is typically applied to single-class or multi-labeled classifiers. This setup is only conceivable because the ground truth is Gaussian. We interpreted the results as the optimizer perceiving the loss function as the weighted sum of the prediction's logarithm, as defined by the equation \ref{eq:FPLoss}, which caused it to boost the scores nearer the human fixation positions. In contrast, when a final softmax activation is utilized, the model turns the scores of the last hidden layers into class probabilities. Due to the ground truth encoding, there is a saturation of the loss when every grid cell is considered, as opposed to only examining the probability of the true class in a one-hot encoding configuration, resulting in the model's poor performance. 

\par Additionally, when comparing the single-task model leveraging panoptic features to the models presented previously, we found, in figure \ref{fig:TFP}, the models with high-level feature maps are able to fixate targets much sooner even though the panoptic features can lead to a similar search accuracy, leading to a smaller TFP-AUC score of 3.259 and a higher probability mismatch of 1.514. In addition, this approach is much less efficient as scanpaths generated through panoptic features travel a much greater distance, as depicted in figure \ref{fig:VGG_distances_fp}, leading to a a scanpath ratio score of only 0.463.

\begin{table*}[t]
\caption{\label{tbl:FP_results} Performance evaluation of several best performing models (rows) based on several Fixation Prediction metrics (columns).}
\begin{center}
\begin{tabular}{l|c|c|c|c}
\multicolumn{1}{c|}{}             & Search                                                                          & \multirow{2}{*}{TFP-AUC  $\uparrow$} & Probability           & Scanpath         \\
\multicolumn{1}{c|}{}             & Accuracy $\uparrow$                                                             &                                      & Mismatch $\downarrow$ & Ratio $\uparrow$ \\ \hline \hline
Human                             & 0.990  & 5.200                                & -                     & 0.862            \\ \hline
High-Level Features - One-hot GT  & 0.650                                                                           & 3.068                                & 1.727                 & 0.753            \\
High-Level Features - Gaussian GT & 0.690                                                                           & 3.413                                & 1.360                 & 0.727            \\
Panoptic Features                 & 0.686                                                                           & 3.259                                & 1.514                 & 0.463            \\
Dual - Architecture A             & 0.719                                                                           & 3.496                                & 1.263                 & 0.808            \\
Dual - Architecture C             & 0.701                                                                           & 3.446                                & 1.320                 & 0.791            \\
IRL                               & N/A                                                                             & 4.509                                & 0.987                 & 0.826            \\
BC-LSTM                           & N/A                                                                             & 1.702                                & 3.497                 & 0.406            \\
Random Scanpath                   & 0.235                                                                           & 1.150                                & 3.858                 & -               
\end{tabular}
\end{center}
\end{table*}

\subsubsection{Target Detection}

\par For the target detection task we used \textbf{accuracy}, \textbf{precision} and \textbf{recall} as metrics. The fine-tuned classifiers with foveation radius of 50 pixels had the maximum performance for all measures, with a mean accuracy, precision and recall of 82.1\%, 86.9\% and 75.4\%, respectively, whereas the configuration of 75 pixels achieved the lowest accuracy and recall. Regarding each individual task, the models with the best performance were those detecting the bottle and stop sign classes, with the first dominating in accuracy and recall with scores of 92.7\% and and 93.2\%, respectively, and the second in precision scoring 100\%, all averaged across fovea settings. Additionally, we may remark that the knife, microwave, and potted plant jobs had the most variation in metrics. The first model, with a fovea of 50 pixels, has a recall score of 80.0\% but an accuracy score of 48.9\%, meaning that the model inaccurately anticipated the existence of the target more than half of the time while rarely reporting positive values as target present. Contrarily, the potted plant model with a fovea of 50 pixels scored 97.9\% for precision and 52.3\% for recall.

\par
\begin{figure}[h]
    \centering
    \begin{subfigure}[b]{0.45\linewidth}
        \includegraphics[width=\textwidth]{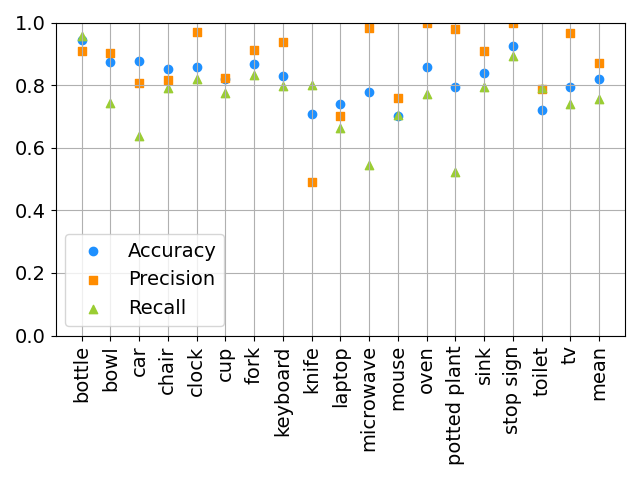}%
        \caption{\label{subfig:det_finetuned_metrics_50} Fine-Tuned Classifier}
    \end{subfigure}
    \qquad
    \begin{subfigure}[b]{0.45\linewidth}
        \includegraphics[width=\textwidth]{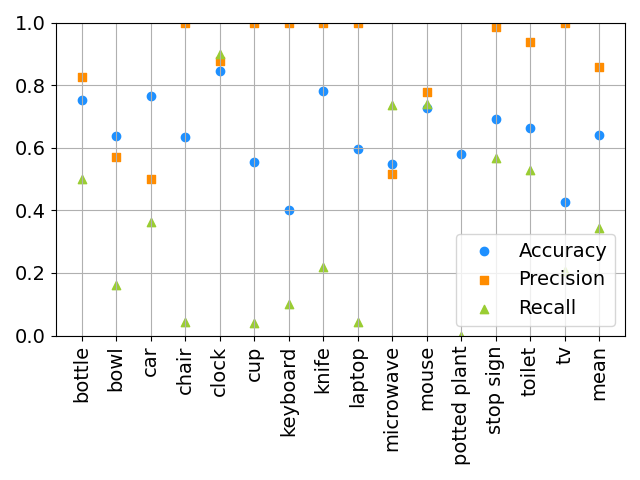}%
        \caption{\label{subfig:det_baseline_metrics_50} Baseline Classifier}
    \end{subfigure}
    \caption{Performances of the fine-tuned and baseline target detectors in terms of accuracy, precision and recall, for the fovea size setting of 50 pixels.}
    \label{fig:det_pretrained_metrics}

\end{figure}

 In turn, the baseline pre-trained model attained incredibly high ranges across metrics for the vast majority of the classes, as seen in figure \ref{subfig:det_baseline_metrics_50}. Although its average precision is similar to the previous models, its recall is incredibly low, averaging the worst in the foveation size with 50 pixels context at 28.1\%. We attribute this to the classes groupings and the differences between the training and testing datasets. We additionally noted a slight increase of the mean scores with the increase of the foveation radius, across all metrics, with its highest accuracy, precision and recall scores being 64.0\%, 85.7\% and 34.3\%, respectively.

\par Additionally, to better comprehend the performance of our model, we compare it to various state-of-the-art benchmarks. The VGG-16 model, trained on the ImageNet dataset, is able to reach a 71.3\% top-1 accuracy when classifying 1,000 classes on the validation split of the same dataset \cite{b34}, which our fine-tuned model outperforms in this problem. This increase in performance is likely attributable to the nature of our task, as we are performing binary classification whereas the benchmark presented for the VGG-16 was performed on multi-class classification of one thousand classes with some classes having highly nuanced visual distinctions, such as mud turtle and box turtle being two distinct classes.  

\par In \cite{b29}, three separate approaches were taken, the first being a K nearest neighbors approach, where the feature maps of each fixation point, during testing, are evaluated against the labeled feature maps obtained from the training dataset of the same task. The prediction is then computed according to the majority label of the closest neighbors. The second approach is a simple method in which a comparison is made between the test sample and the feature maps averaged across the training data. This comparison is made through a loss, which, in the cases it falls bellow a pre-defined threshold, detects the target as present. Lastly, the same approach as our baseline was taken. These three achieved accuracies of 77.8\%, 30.5\% and 70.24\%, respectively. The first two results may be attributable, once more, to the lower size of the dataset used in their study, while the last model outperforms our baseline probably due to a better match between class equivalencies, resulting in a higher degree of similarity between the training and testing datasets.

\subsection{Dual Task}
\par
\begin{figure}[h]
    \centering
    \begin{subfigure}[b]{0.45\linewidth}
        \includegraphics[width=\textwidth]{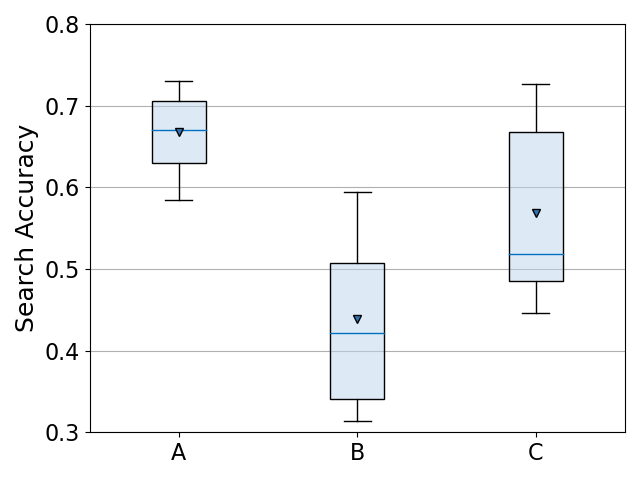}%
        \caption{Architecture.}
        \label{subfig:dual_arch_fix}%
    \end{subfigure}
    \qquad
    \begin{subfigure}[b]{0.45\linewidth}
        \includegraphics[width=\textwidth]{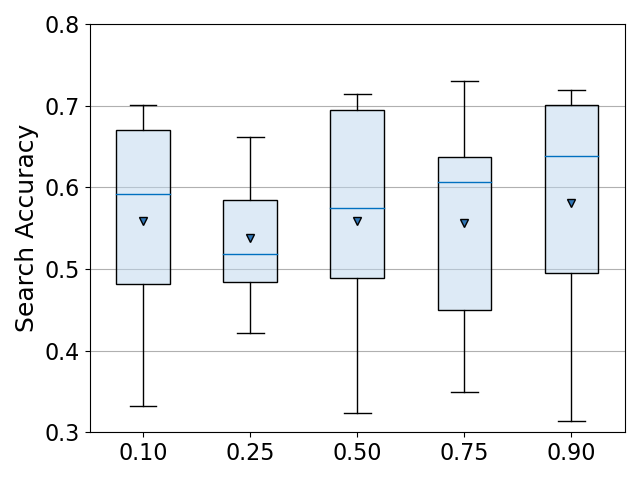}%
        \caption{Fixation loss' weight.}
        \label{subfig:dual_weight_fix}%
    \end{subfigure}
    \vskip \baselineskip
    \begin{subfigure}[b]{0.45\linewidth}
        \includegraphics[width=\textwidth]{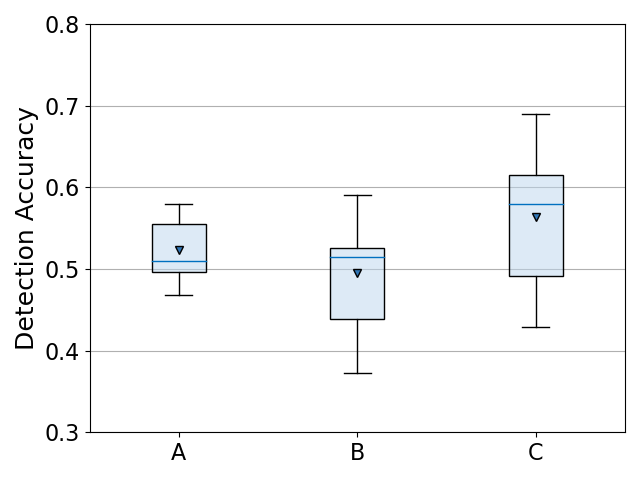}%
        \caption{Architecture.}
        \label{subfig:dual_arch_det}%
    \end{subfigure}
    \qquad
    \begin{subfigure}[b]{0.45\linewidth}
        \includegraphics[width=\textwidth]{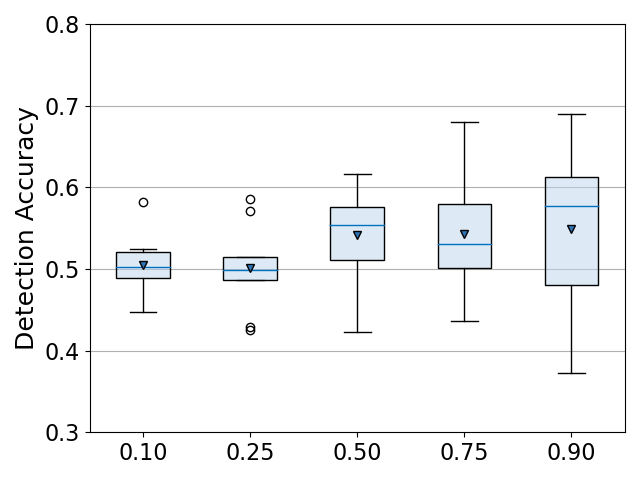}%
        \caption{Fixation loss' weight.}
        \label{subfig:dual_weight_det}%
    \end{subfigure}
\caption{\label{fig:Dual_boxplot_fix} Box-plots of the Search accuracy, on the top, and Detection accuracy, on the bottom, grouped by their configuration, with the mean values and outliers depicted as triangles and circles, respectively.}
\end{figure}

 Finally, we evaluate the dual task models in their performances in both tasks. In figure \ref{subfig:dual_arch_fix}, we can see that architecture B fared the worst, and, architecture A outperforms more than half of the models with architecture C, in addition to achieving the highest search accuracy of 73.4\% when configured with a fovea size of 100 pixels and a fixation loss' weight of 0.75. 

\par In terms of its detection performance, the dual task model with architecture C, a fovea size of 50, and $w_{fix}$ set to 0.90 achieved the highest target presence detection rate of 68.7\%. In figure \ref{subfig:dual_arch_det}, considering the quartiles and upper limit of its performance, the architecture of design C is deemed to be the most effective. Regarding the weight of the fixation loss, we can also see that models trained with bigger values obtained a larger interquartile range than models trained with smaller values. 

\par In addition, we noted that the best scanpath prediction model only achieved a detection accuracy of 49.7\% while the best target presence predictor only achieved a search accuracy of 63.9\%. In light of this, we also considered the average of both of these metrics to determine which models were more capable of performing both tasks simultaneously. The majority of the time, design A earned a higher score than design C, while design B ranked the lowest. Regarding the remaining parameters, we can only observe that a bigger fovea radius led to higher average scores, and a higher fixation loss' weight resulted in a better top score, with the exception of setting $w_fix = 0.25$. The model configured with architecture C, a fovea size of 75 pixels, and $w fix$ set to 0.90 achieved the top score of 67.7\% with search and detection accuracies of 70.1\% and 65.3\%, respectively.

\par
\begin{figure}[h]
    \centering
    \begin{subfigure}[b]{0.95\linewidth}
        \includegraphics[width=\textwidth]{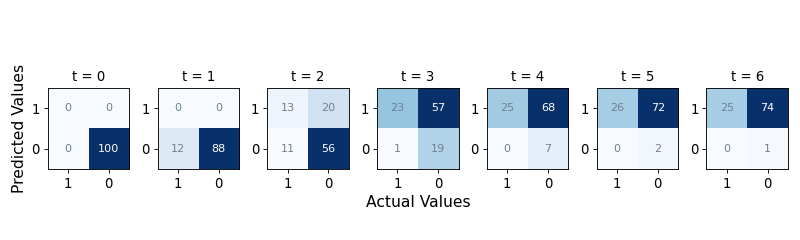}%
        \label{subfig:dual_confusion_fifirst}%
        \caption{Architecture A.}
    \end{subfigure}
    \vskip \baselineskip
    \begin{subfigure}[b]{0.95\linewidth}
        \includegraphics[width=\textwidth]{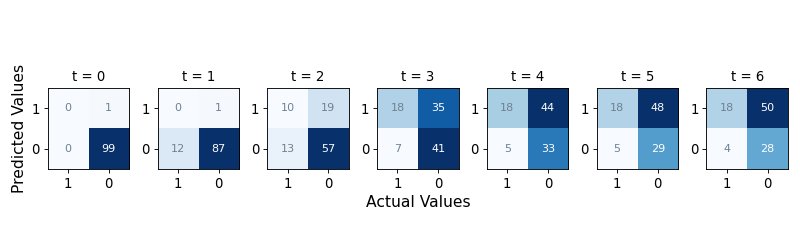}%
        \label{subfig:dual_confusion_concat}%
        \caption{Architecture C.}
    \end{subfigure}
    
    \caption{Target detection confusion matrices, in percentage, separated by time step for the best overall models with fixation first, on top, and concatenation, bellow, architectures. }
    \label{fig:dual_confusion}

\end{figure}

 During testing, we  discovered a trend in the target detection predictions from models with architecture A, which is exemplified in figure \ref{fig:dual_confusion}, which plots the confusion matrices of the temporal predictions of the models with the highest overall scores for architectures A and C. During the second half of the gaze sequence, the former model hardly predicts that the target is absent, indicating that the recurrent nature of the detection branch biases the model to output the pattern learned from the training data that, for target present images, the object is detected a vast majority of the time, as stated previously. In contrast, we can see that the model with architecture C seeks to differentiate between target present and target missing fixations at later time steps.

\par  Finally, when comparing the fixation prediction performance of this approach to that of previously presented models, the dual approach led to a higher search accuracy, as seen in figure \ref{tbl:FP_results}, resulting in a higher TFP-AUC score of 3.496 and 3.446 and a lower probability mismatch of 1.263 and 1.320 for the overall best models with architectures A and C, respectively. In addition, the two models exhibit a higher search efficiency with scanpath ratios of 0.808 and 0.791, respectively, because, as shown in \ref{fig:VGG_distances_fp}, the model with architecture C travels a shorter distance during the first time step and the model with architecture A is able to maintain an efficient search throughout the remainder of the scanpath.

\section{Conclusions}

\par In this thesis, we described two methods for predicting the presence or absence of a target in an image with foveated context. In the first, a two-stage pipeline system, the fixation prediction module produced the best results while using high-level feature maps as image representations, both in terms of search accuracy and search efficiency. In addition, we discovered that the usage of a Gaussian ground truth label encoding, which is our first contribution, enhanced search accuracy. This novel representation captures the spatial structure of the problem by not only encouraging predictions to the exact discretized human fixation positions, but also favoring attempts to cells near these locations.

\par In the fixation prediction models with panoptic features, a cumulative mask representation and a final sigmoid activation function were shown to be advantageous. The latter is only conceivable when using a Gaussian ground truth label. This setup imposes that the categorical cross entropy loss performs a weighted sum of the loss contribution of each cell, where positions closer to the correct location are assigned a higher weight. However, while these image representations achieved similar search accuracy to the high-level feature maps, they presented a lower search efficiency. This suggests that the ConvLSTM is better suited for high-level feature inputs, but the IRL learning strategy was able to harness the full potential of the panoptic features by predicting both the action and state representations at each time step. 

\par Two classifiers performed the target presence prediction portion of the two-stage pipeline model. A pre-trained VGG-16 was used as a baseline for the dual task model, and a fine-tuning approach set the performance ceilings for the detection task.  

\par The final contribution of our thesis is a dual-task model that executes both tasks concurrently while enabling information sharing between them by executing a common input transformation and establishing linking channels throughout each task branch. This multi-task approach improved search precision when the task prediction branch initiated the predictions, i.e. in designs A and C. However, the former suffered a reduction in detection accuracy, whilst the latter achieved the maximum score when compared to our baseline method. We additionally concluded that the proposed approach of employing a ConvLSTM layer during the detection branch resulted in the model basing its predictions on the temporal pattern of target detection, as opposed to utilizing the input visual features as the basis for its prediction. 

\section{Future Work}

\par  The gap in search accuracy performance between the human performance ceiling and our best results reflects the challenge of duplicating task-guided attention at the start of the scanpath. As the model's target fixation reaches a plateau earlier than its training data, future research should concentrate on increasing the fixation prediction of the first time-steps and not so much the last. In the testing phase, it is important to test a new fixation prediction algorithm that gives greater weight to sequences with higher scores at the beginning of the scanpath. 

\par Regarding the target's presence prediction during the dual-task approach, we have demonstrated that the use of a recurrent layer biases the model towards temporal patterns of target detection. An alternative solution would be for the task branch to generate the input image simultaneously, simulating an encoder-decoder, so as to require the model to maintain its knowledge of high-level features in its hidden states. 

\par Future research should also investigate a visual transformer-based design, as it has shown promising results in similar image classification and goal-directed search tasks.

\end{document}